\documentclass[3p]{elsarticle}

\usepackage{hyperref}
\usepackage{caption}
\usepackage{amsmath}
\usepackage{subfigure}

\usepackage{algorithm}
\usepackage{algorithmic}
\usepackage{booktabs}
\usepackage{threeparttable}
\usepackage{bm}
\usepackage{multirow}
\usepackage{indentfirst}
\usepackage{adjustbox}
\usepackage{colortbl}  %彩色表格需要加载的宏包
\usepackage{xcolor}
\usepackage{amssymb}

\journal{Journal of \LaTeX\ Templates}

\begin{document}
\captionsetup[figure]{labelfont={bf},labelformat={default},labelsep=period,name={Fig.}}

\begin{frontmatter}

\title{IELDG: Suppressing Domain-Specific Noise with Inverse Evolution Layers for Domain Generalized Semantic Segmentation}

%% Group authors per affiliation in footnotes:
\author[mymainaddress]{Qizhe Fan}

\author[mysecondaryaddress]{Chaoyu Liu}

\author[mythirdaryaddress]{Zhonghua Qiao}

\author[mymainaddress]{Xiaoqin Shen\corref{mycorrespondingauthor}}
\cortext[mycorrespondingauthor]{Corresponding author}
\ead{xqshen@xaut.edu.cn}

\address[mymainaddress]{School of Sciences, Xi’an University of Technology, Xi’an 710054, China}
\address[mysecondaryaddress]{Department of Applied Mathematics and Theoretical Physics, University of Cambridge, UK}
\address[mythirdaryaddress]{Department of Applied Mathematics, The Hong Kong Polytechnic University, Hung Hom, Hong Kong}

\begin{abstract}
Domain Generalized Semantic Segmentation (DGSS) focuses on training a model using labeled data from a source domain, with the goal of achieving robust generalization to unseen target domains during inference. A common approach to improve generalization is to augment the source domain with synthetic data generated by diffusion models (DMs). However, the generated images often contain structural or semantic defects due to training imperfections. Training segmentation models with such flawed data can lead to performance degradation and error accumulation. To address this issue, we propose to integrate inverse evolution layers (IELs) into the generative process. IELs are designed to highlight spatial discontinuities and semantic inconsistencies using Laplacian-based priors, enabling more effective filtering of undesirable generative patterns. Based on this mechanism, we introduce IELDM, an enhanced diffusion-based data augmentation framework that can produce higher-quality images. Furthermore, we observe that the defect-suppression capability of IELs can also benefit the segmentation network by suppressing artifact propagation. Based on this insight, we embed IELs into the decoder of the DGSS model and propose IELFormer to strengthen generalization capability in cross-domain scenarios. To further strengthen the model's semantic consistency across scales, IELFormer incorporates a multi-scale frequency fusion (MFF) module, which performs frequency-domain analysis to achieve structured integration of multi-resolution features, thereby improving cross-scale coherence. Extensive experiments on benchmark datasets demonstrate that our approach achieves superior generalization performance compared to existing methods.
\end{abstract}


\end{frontmatter}

%\linenumbers

\section{Introduction}\label{section:1}

\begin{figure}[!t]
\centering
\includegraphics[width=3.2in]{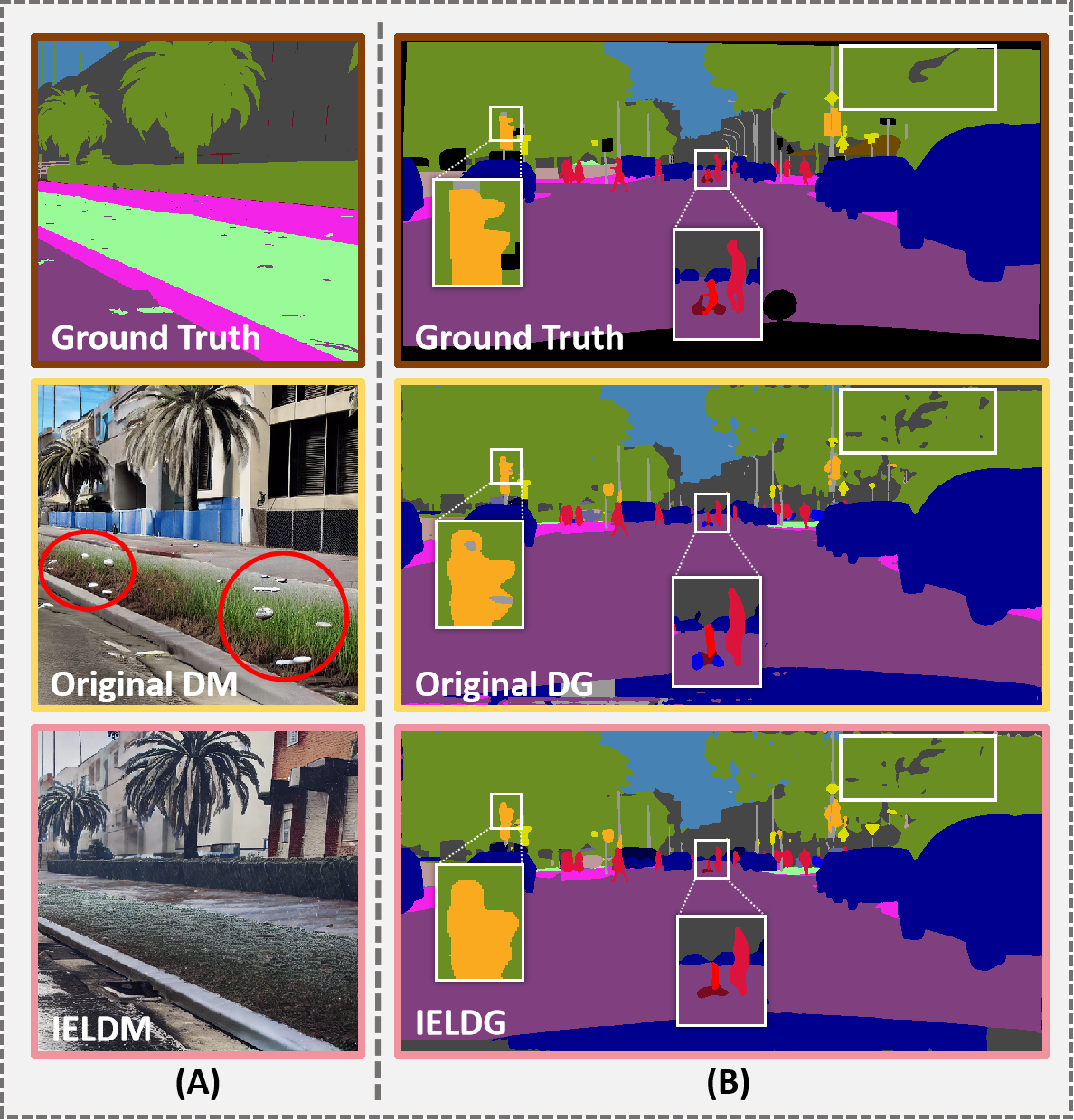}
\caption{\textbf{(A) Comparison of synthetic images generated by original DM and our IELDM.} The original DM produces images with noticeable artifacts and semantic inaccuracies (e.g., \emph{vegetation} incorrectly rendered as \emph{fence}), which may mislead the DGSS model during training and hinder its generalization capability. In contrast, images generated by IELDM exhibit improved structural integrity and semantic correctness, providing higher-quality supervision for model training.  \textbf{(B) Comparison of segmentation predictions produced by original DG and our IELDG.} The original DG results contain evident segmentation flaws and misclassified regions, whereas IELDG yields more accurate and spatially coherent predictions, demonstrating superior generalization to unseen domains.}
\label{figure:intro}
\end{figure}

Deep neural networks (DNNs) have achieved remarkable success in dense prediction tasks such as semantic segmentation \cite{chen2017deeplab,xie2021segformer,cheng2022masked}, particularly when the training and testing data share the same distribution. However, this dependence on distributional consistency presents significant challenges in real-world applications, where domain shifts \cite{quinonero2008dataset} commonly arise due to variations in illumination, weather, sensor characteristics, or scene layouts \cite{ai2024domain,chen2024dual,yan2023deep}. Such distributional discrepancies can severely impair the generalization ability of segmentation models, thereby resulting in substantial performance degradation on unseen target domains \cite{cordts2016cityscapes,yu2020bdd100k,neuhold2017mapillary}. To address this issue, domain adaptive semantic segmentation (DASS) has attracted increasing attention \cite{tsai2018learning,tranheden2021dacs,zhang2021prototypical,hoyer2022daformer,hoyer2022hrda,hoyer2023mic}. DASS methods aim to bridge the domain gap by aligning feature distributions between a labeled source domain and an unlabeled but accessible target domain. Although effective in many scenarios, these approaches typically rely on access to target domain data during training, limiting their practicality in dynamic or privacy-sensitive environments. To overcome these limitations, domain generalized semantic segmentation (DGSS) has emerged as a promising alternative \cite{pak2024textual,benigmim2024collaborating,wei2024stronger,bi2024learning,zhang2025mamba,yun2025soma,bi2025nightadapter}. DGSS seeks to learn domain-invariant representations using only source domain data, without requiring any exposure to target domains during training. By decoupling model training from target domain dependence, DGSS provides a more robust and scalable solution to domain shifts, thereby offering enhanced generalization in complex and diverse real-world settings.

Recently, the rapid development of diffusion models (DMs) \cite{ho2020denoising,rombach2022high,zhang2023adding} has provided new opportunities for advancing DGSS \cite{jia2024dginstyle,benigmim2024collaborating,niemeijer2024generalization}. Owing to their powerful generative capacity and ability to model complex data distributions, DMs have been increasingly adopted to synthesize diverse and style-rich images that simulate target domain variations. Such synthetic data can be used to augment the source domain and improve model robustness under domain shifts. However, despite their potential, images generated by diffusion-based frameworks often exhibit structural inconsistencies or semantic defects (second row of Fig. \ref{figure:intro} (A)) due to imperfect training or insufficient domain constraints. Training DGSS models directly on such flawed data may introduce noise into the learning process, hinder feature alignment, and ultimately degrade generalization performance. To address this challenge, we propose a novel image generation framework, \textbf{IELDM}, which integrates inverse evolution layers (IELs) \cite{liu2025iels} with DM to enhance the realism and structural fidelity of the generated images. IELs employ Laplacian-based filtering to deliberately amplify subtle structural defects in generated images, thereby encouraging the neural network to minimize such defects during generation. By integrating this mechanism into the generative process, IELs provide explicit gradient feedback that encourages the model to refine low-quality regions, resulting in outputs that are both semantically coherent and structurally faithful (third row of Fig. \ref{figure:intro} (A)). This enhanced data quality enables the DGSS model to learn more robust domain-invariant features, thereby improving its generalization performance on unseen target domains.

In addition, it is noteworthy that recent advances in DGSS have predominantly concentrated on enhancing generalization capabilities by either enforcing domain-invariant feature learning \cite{pan2018two,ahn2024style,jing2023order,wu2022siamdoge,zhao2022style,choi2021robustnet,peng2022semantic,niu2025exploring} or introducing style diversification strategies \cite{zhong2022adversarial,chattopadhyay2023pasta,peng2021global,ho2020denoising,rombach2022high,zhang2023adding,podell2023sdxl}. While these approaches have demonstrated effectiveness, they may still face limitations under significant domain shifts, where the reliability of model predictions tends to degrade due to increased semantic ambiguity and domain discrepancy. When such suboptimal predictions are incorporated during training as supervisory signals, they can introduce semantic noise and reinforce incorrect patterns, ultimately hindering generalization, as evidenced by the second row of Fig. \ref{figure:intro} (B). In fact, prediction defects are inherent to any model and cannot be entirely eliminated. Nevertheless, devising effective strategies to mitigate these imperfections during training can markedly enhance overall model performance. Accordingly, inspired by the proven effectiveness of IELs in IELDM, we also adapt them into the DGSS pipeline to mitigate such defects and enhance the model’s robustness. Specifically, IELs are tailored to identify and accentuate uncertain or erroneous regions within the predicted segmentation maps, thereby facilitating the model’s ability to recognize and rectify its own weaknesses during training, as exemplified in the third row of Fig.~\ref{figure:intro} (B). In detail, we embed IELs at four hierarchical feature levels within the pixel decoder of Mask2Former \cite{cheng2022masked}, placing them immediately after each intermediate-resolution feature map. By integrating IELs across multiple scales, the model is encouraged to iteratively suppress prediction artifacts that may emerge at different semantic depths. We refer to this enhanced architecture as \textbf{IELFormer}.

To promote semantic consistency across feature resolutions in IELFormer, we propose a \textbf{multi-scale frequency fusion (MFF)} module, which performs frequency-domain decomposition via fast Fourier transform (FFT) to structurally integrate multi-scale features. Specifically, each feature map is transformed into its frequency spectrum, where it is factorized into amplitude and phase components. The amplitude encodes the magnitude of semantic activations, reflecting the spatial distribution of feature intensity, while the phase preserves fine-grained structural details and contour information critical for maintaining spatial coherence. By jointly leveraging these complementary components, the MFF module enables more semantically aligned and structurally consistent feature fusion across scales.

By integrating the high-quality data generated by the proposed IELDM and the structurally guided IELFormer architecture, which incorporates the MFF module as an essential component for enhancing multi-scale feature fusion, we establish a unified and robust DGSS framework, termed \textbf{IELDG}. Through this comprehensive design, the model demonstrates improved robustness under domain shifts and enhanced generalization to unseen environments. Building upon this foundation, we highlight the main contributions of our IELDG as follows:
\begin{itemize}
\item[$\bullet$] We propose IELDM, a novel diffusion-based image generation framework enhanced by inverse evolution layers, which improves the structural and semantic fidelity of synthetic training data for DGSS.
\item[$\bullet$] We introduce IELFormer, an enhanced segmentation architecture that integrates IELs at multiple semantic scales to explicitly highlight and suppress prediction artifacts during training.
\item[$\bullet$] We design a multi-scale frequency fusion (MFF) module that performs frequency-domain decomposition and integration of multi-resolution features, promoting cross-scale semantic consistency and spatial alignment.
\item[$\bullet$] Extensive experiments on widely-used DGSS benchmarks demonstrate that the proposed IELDG framework consistently outperforms previous state-of-the-art methods under challenging domain shift conditions.
\end{itemize}

%%%%%%%%%%%%%%%%%%%%%%%%%%%%%%%%%%%%%%%%%%%%%%%%%%%%%%%%%%%%%%%%%%%%%%%%%%%%%%%%%%%%%%%%%%
%%%%%%%%%%%%%%%%%%%%%%%%%%%%%%%%%%%%%%%%%%%%%%%%%%%%%%%%%%%%%%%%%%%%%%%%%%%%%%%%%%%%%%%%%%

\section{Related works}\label{sec:Related works}

\subsection{Domain Generalized Semantic Segmentation (DGSS)}\label{sec:2.1 DGSS}

DGSS is dedicated to improving the generalization capability of segmentation models. It entails training networks on one or more source domains with the aim of empowering them to perform accurately on entirely unseen target domains. Existing DGSS approaches can be broadly divided into two main categories: (i) data-centric strategies, which enrich the training distribution through augmentation or style variation to simulate domain shifts, and (ii) feature-centric methods, which aim to learn domain-invariant representations by mitigating domain-specific discrepancies.

Methods based on the first category typically apply diverse data augmentation strategies at the input level to simulate domain shifts. For instance, PASTA \cite{chattopadhyay2023pasta} generates augmented views by applying structured perturbations to the amplitude spectra in the Fourier domain, emphasizing changes in high-frequency components over low-frequency ones. AdvStyle \cite{zhong2022adversarial} treats the style feature as a learnable parameter, which is adversarially optimized to generate robust training images. GLTR \cite{peng2021global} introduces global and local texture randomization to diversify source textures and reduce texture bias, with a consistency regularization to align both mechanisms. In recent studies, the remarkable performance of diffusion models \cite{ho2020denoising,rombach2022high,zhang2023adding,podell2023sdxl} in image synthesis has led to their adoption for generating data \cite{jia2024dginstyle,benigmim2024collaborating,niemeijer2024generalization} that closely resemble the target domain, thereby effectively enriching domain diversity and enhancing the representational capacity of learned features. Furthermore, to facilitate more flexible augmentation, several studies \cite{fahes2024simple,fan2023towards,lee2022wildnet,vidit2023clip,udupa2024mrfp} have explored feature stylization techniques to improve model robustness against domain shifts. In contrast, alignment-based approaches mitigate domain-specific discrepancies by leveraging feature normalization \cite{pan2018two}, whitening strategies \cite{choi2021robustnet}, or incorporating regularization objectives to minimize inconsistencies induced by synthetic domain shifts \cite{ahn2024style,jing2023order,wu2022siamdoge,zhao2022style}.

While existing methods have achieved promising results, the inevitable presence of prediction biases during training remains insufficiently addressed. When such biases are used as supervisory signals, they can gradually propagate and undermine generalization. Moreover, data-centric approaches based on diffusion models often produce images with structural or semantic defects, further degrading model performance. To address these issues, we introduce IELs \cite{liu2025iels} into both DM and DGSS, where they respectively enhance the structural fidelity of generated images and refine less reliable predictions, jointly improving robustness to domain shifts.

\subsection{Diffusion Models (DMs) for DGSS}\label{sec:2.2 DM}

Diffusion models (DMs), a prominent class of generative models \cite{rombach2022high,brown2020language,touvron2023llama,guo2025deepseek}, have demonstrated exceptional capability in synthesizing photorealistic images \cite{niemeijer2024generalization,jia2024dginstyle,benigmim2023one,sariyildiz2023fake}, thereby facilitating high-quality dataset generation and significantly advancing performance across various vision tasks. Leveraging the powerful generative capability of DMs, recent DGSS studies have increasingly explored their use to enhance model adaptability to unseen domains. CLOUDS \cite{benigmim2024collaborating} employs DM to diversify generated content, effectively capturing a wide range of variations within the potential target domain distribution. DGInStyle \cite{jia2024dginstyle} tackles the problem of domain-specific semantic guidance in image synthesis by adapting a pretrained latent DM through a novel style swap strategy, which effectively incorporates semantic control into the model’s expressive generative capacity. DIDEX \cite{niemeijer2024generalization} introduces a diffusion-driven domain extension framework that generates a pseudo-target domain via diverse textual prompts, allowing precise control over style and content while promoting high diversity to support effective model generalization.
PTDiffSeg \cite{gong2023prompting} introduces a scene prompt with randomized augmentation to enhance the disentanglement of domain-invariant representations during training, and further employs an unsupervised prompt learning mechanism for efficient test-time adaptation to target domains.

Although DM-based approaches have shown promise, the generated images may still exhibit structural defects or semantic inconsistencies, introducing noise and bias that hinder generalization. To mitigate this, we propose IELDM, which integrates IELs \cite{liu2025iels} into DM to enhance structural fidelity and semantic consistency, enabling more reliable pseudo-domain construction and improving robustness in downstream segmentation.

%%%%%%%%%%%%%%%%%%%%%%%%%%%%%%%%%%%%%%%%%%%%%%%%%%%%%%%%%%%%%%%%%%%%%%%%%%%%%%%%%%%%%%%%%%
%%%%%%%%%%%%%%%%%%%%%%%%%%%%%%%%%%%%%%%%%%%%%%%%%%%%%%%%%%%%%%%%%%%%%%%%%%%%%%%%%%%%%%%%%%

\section{Methods}\label{sec:methods}

The objective of Domain Generalized Semantic Segmentation (DGSS) is to develop a segmentation model $\mathcal{M}$ using training data from a source domain $\mathcal{S}$, enabling it to generalize effectively to arbitrary unseen target domains $\mathcal{T}$ with distinct data distributions. Let $\mathcal{D}_S = \{(\mathbf{x}^{S}_i, \mathbf{y}^{S}_i)\}_{i=1}^{N^S}$ with $N^S$ samples denote the labeled source domain data and $\mathcal{D}_G = \{(\mathbf{x}^{G}_i, \mathbf{y}^{G}_i)\}_{i=1}^{N^G}$ with $N^G$ samples represent the generated labeled dataset, where $\mathbf{x}_i$ denotes an RGB source image and $\mathbf{y}_i$ is its associated one-hot encoded ground-truth label. Given a pre-trained backbone network $f_\varphi$ with parameters $\varphi$ and a decoder head $h_\theta$ parameterized by $\theta$, the optimization objective for DGSS can be formulated as:

\begin{equation}\label{eq:dgss optimization objective}
\underset{\varphi,\theta}{\arg\min} \sum_{i=1}^{N} \mathcal{L}_{DGSS}\left[h_{\theta}\left(f_\varphi(\mathbf{x}^{S}_{i}\oplus\mathbf{x}^{G}_{i})\right), (\mathbf{y}^{S}_{i}\oplus\mathbf{y}^{G}_{i})\right],
\end{equation}
where $\mathbf{x}^{S}_{i}\oplus\mathbf{x}^{G}_{i}$ indicates alternately feeding both original and generated source images into the network, and $N$ represents the total number of input samples.
\\
\\
\noindent \textbf{Overview of IELDG:} IELDG incorporates inverse evolution layers (IELs) into neural networks, utilizing them as negative property amplifiers to regularize the network behavior. By amplifying certain undesirable characteristics, the approach compels the network to produce more favorable results (Sec. \ref{sec:3.1 IELs}). To enhance content diversity in the source domain while maintaining DGSS model prediction accuracy, we integrate IELs with a conditional text-to-image diffusion model (DM). This synergistic combination generates higher-quality synthetic images for domain-expanded training, achieving improved generalization without compromising segmentation precision (Sec. \ref{sec:3.2 IELDM}). Furthermore, we incorporate IELs into the DGSS decoder to mitigate prediction artifacts and boost segmentation accuracy. Simultaneously, we employ a multi-scale frequency fusion (MFF) strategy prior to IELs to strengthen feature representation (Sec. \ref{sec:3.3 IELFormer}). The strategic integration of these approaches substantially improves generalization capabilities across diverse domains.

\subsection{Theoretical Formulation of Inverse Evolution Layers (IELs)} \label{sec:3.1 IELs}

\begin{figure}[!t]
\centering
\includegraphics[width=3.6in]{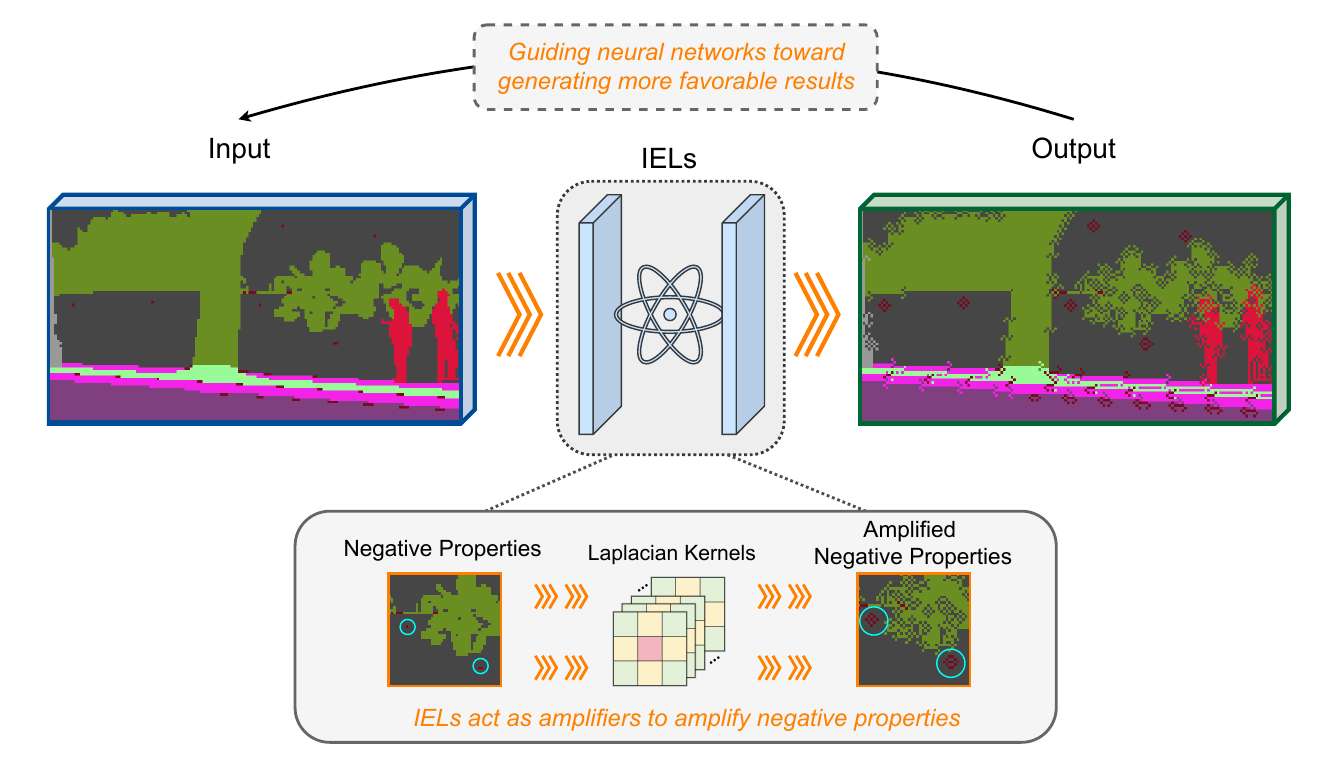}
\caption{\textbf{Core Mechanism of IELs.} IELs operate as intelligent negative property amplifiers through their Laplacian kernel-based architecture, which systematically detects and exacerbates undesirable characteristics in neural network outputs. When suboptimal features (e.g., noise artifacts or geometric irregularities) are fed into IELs, the architecture performs spectral decomposition to quantify these deficiencies, subsequently applying controlled amplification to generate adversarial feedback signals. This process creates a self-regulating optimization loop where the network is compelled to develop compensatory representations that inherently resist the amplified artifacts while preserving semantic fidelity.}
\label{figure:IEL_structure}
\end{figure}

Currently, most existing computer vision models either inadequately address or completely overlook the negative properties that emerge during model training. These unmitigated negative characteristics, particularly gradient instability, feature entanglement, and prediction bias, significantly compromise model robustness and degrade inference accuracy. This fundamental shortcoming in current architectures reveals an urgent demand for structured solutions capable of identifying and mitigating adverse training dynamics.

In this context, inverse evolution layers (IELs)\cite{liu2025iels} emerge as a pivotal solution for model regularization and quality enhancement (Fig. \ref{figure:IEL_structure}). This architectural innovation establishes a new paradigm by systematically integrating with neural networks to function as intelligent negative property amplifiers. The underlying mechanism works by deliberately magnifying specific undesirable characteristics during training, thereby forcing the network to develop more effective countermeasures and ultimately yield superior, more stable results. Through this self-regulating process, IELs successfully mitigate persistent challenges associated with uncontrolled negative characteristics. \textbf{It is important to note that IELs are applied exclusively during the training phase and are removed during inference.} For comprehensive technical details regarding IELs' implementation and rigorous theoretical proofs of their effectiveness, we direct readers to the foundational reference \cite{liu2025iels}, which provides exhaustive mathematical analysis and experimental validations.

\subsection{IELDM: Integration of IELs with Diffusion Model (DM)} \label{sec:3.2 IELDM}

\begin{figure*}[!t]
\centering
\includegraphics[width=6.5in]{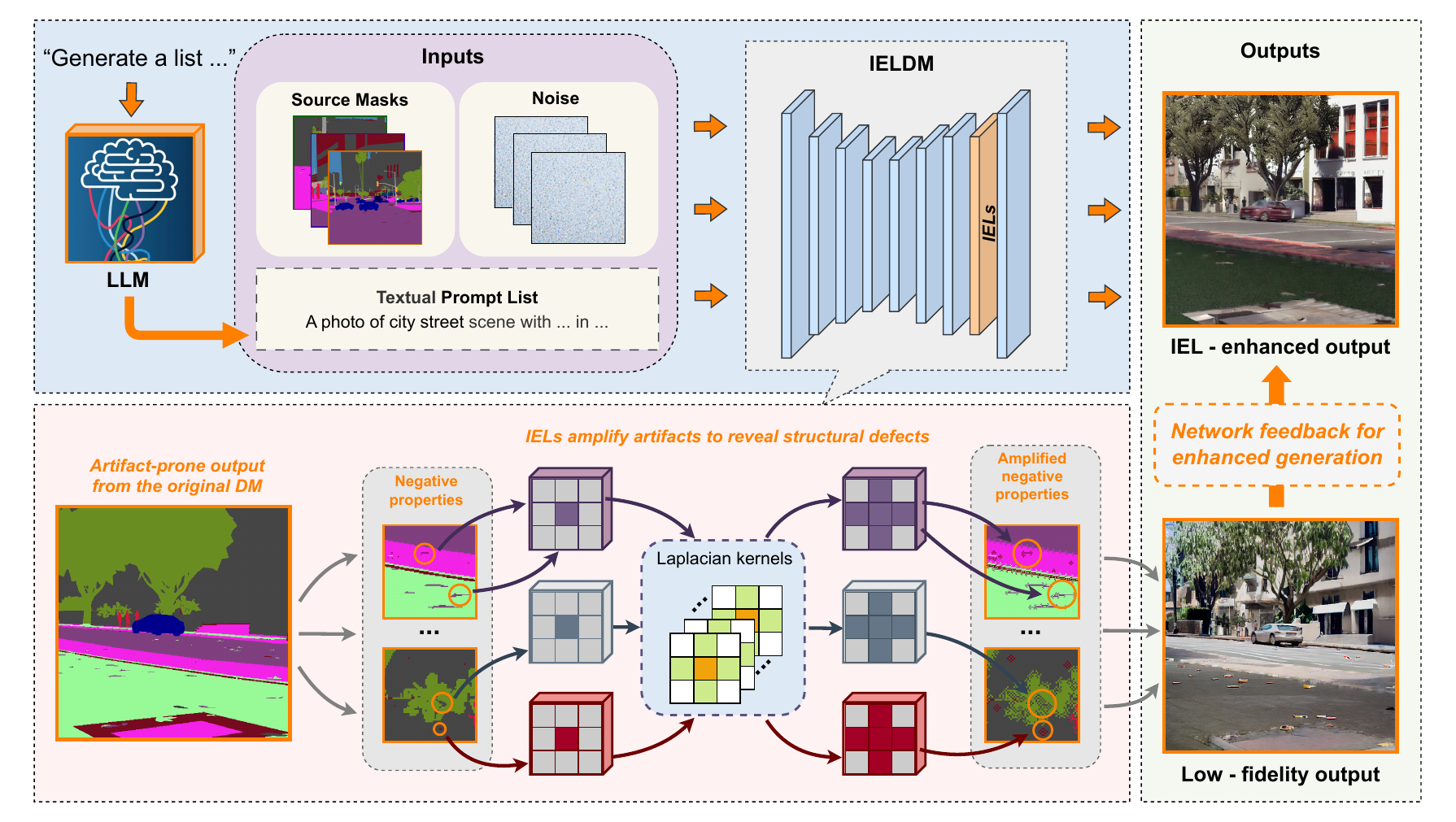}
\caption{\textbf{Overview of the IELDM framework for enhanced image generation.} The proposed IELDM framework integrates IELs into a diffusion-based image generation pipeline to improve structural fidelity and semantic consistency. Given a semantic mask, random noise, and a textual prompt list generated by a large language model (LLM), the framework synthesizes realistic street scenes. The lower portion depicts the core mechanism of IELs: when outputs from the original DM manifest undesirable characteristics, the Laplacian kernels within IELs will capture the underlying deficiency patterns and subsequently amplify them. The amplified artifacts are then fed back into the network as a corrective signal, enabling the generator to iteratively refine its outputs. This feedback mechanism encourages the model to produce more visually coherent and semantically accurate images.}
\label{figure:IELDM}
\end{figure*}

The rapid advancement of image generation models has significantly impacted a wide range of fields, including computer vision \cite{qiu2024aligndiff,namekata2024emerdiff,couairon2024diffcut,tian2024diffuse}, natural language processing \cite{liu2024text,zhang2023diffusum,lovelace2024diffusion,fang2025llama}, and medical image processing \cite{zhang2024diffboost,wu2024medsegdiff}. In domain generalized semantic segmentation (DGSS), a growing body of research has employed image generation models to augment source-domain datasets \cite{jia2024dginstyle,benigmim2024collaborating,niemeijer2024generalization}, aiming to enhance the model’s generalization capability to unseen target domains. Despite their potential, the quality of these synthetic images has often been underexplored. In practice, low-fidelity or artifact-laden generated samples can introduce undesirable noise into the training process, thereby impairing feature learning, diminishing segmentation accuracy, and even causing negative transfer. To mitigate these risks, we propose incorporating IELs into the image generation pipeline, with the objective of improving both the perceptual fidelity and structural integrity of the synthesized images, ultimately providing higher-quality data for robust DGSS model training.
\\

\noindent \textbf{Mask conditioned diffusion model (DM).} To achieve high content diversity in the synthetic data, we employ a pretrained DM conditioned jointly on textual descriptions and semantic masks. Specifically, we utilize a ControlNet \cite{zhang2023adding} architecture, where one-hot encoded semantic masks are fed into the control branch to guide the spatial layout of the generated images. The model is supervised using corresponding source domain images to ensure structural consistency and semantic alignment. In addition, we provide textual prompts that describe the overall scene context, including object attributes, environmental conditions, and stylistic elements, in order to control the appearance and semantic variations of the generated images. During the generation process, the semantic mask constrains the spatial arrangement of scene components, while the textual prompt modulates fine-grained visual characteristics. This dual conditioning mechanism enables DM to synthesize diverse and semantically coherent images.
\\

\noindent \textbf{IEL-enhanced DM.} Experimental results (detailed later in Sec. \ref{sec:4.3 Ablation Study on the Number of IELs}) demonstrate that using the aforementioned method alone to augment the source domain dataset often leads to suboptimal outcomes. In certain instances, the generated images exhibit inadequate semantic alignment, leading to inaccurate class representations that can misguide the training process. In other cases, the synthesized outputs contain structural artifacts or visual degradations, introducing undesirable noise into the training pipeline and ultimately impairing model performance. To address these limitations, we incorporate IELs into the generation framework to enhance the structural fidelity and visual quality of the synthesized images.

As illustrated in Fig. \ref{figure:IELDM}, when the intermediate features produced by the original diffusion-based framework exhibit undesirable characteristics, IELs employ their core component, an ensemble of Laplacian kernels, to accentuate and amplify these deficiencies. By applying it to the generated intermediate features, IELs selectively amplify salient undesirable traits that serve as indicators of underlying structural or semantic distortions. This targeted enhancement enables the network to explicitly attend to regions deviating from expected structural patterns. The enhanced signals are then propagated back through the network, serving as a refined feedback mechanism to guide feature refinement and promote more robust representation learning. Unlike traditional loss-based corrections that may overlook subtle inconsistencies, the IEL-driven feedback directly encodes visually and semantically meaningful discrepancies. This not only facilitates more informed gradient updates but also encourages the generator to iteratively correct its outputs in a self-regularizing manner. As a result, the model progressively learns to produce images with improved structural fidelity and enhanced semantic consistency.
\\

\noindent \textbf{LLM for enhancing prompt diversity.} To obtain diverse and semantically rich textual prompts for image generation, we leverage the large language model (LLM) DeepSeek-R1 \cite{guo2025deepseek} to automatically generate scene descriptions conditioned on predefined semantic categories and contextual information. Specifically, we compile the class names corresponding to all semantic masks used for image generation into a list, ensuring a one-to-one correspondence between each set of class names and its associated mask. To further enhance the semantic compositionality of the generated scenes, we diversify the textual descriptions by varying environmental attributes such as lighting conditions, weather, and time of day. Consequently, the prompts generated by LLM are structured in the form: ``\emph{A photo of a city street scene with $\mathbf{C}$ in $\mathbf{X}$}."  To obtain such prompts, we instruct LLM with the following formulation:\emph{ For each set of semantic categories in the predefined list, generate a corresponding textual prompt to be used for generating synthetic images via an image generation model. Each prompt is required to strictly follow the template:``A photo of a city street scene with $\mathbf{C}$ in $\mathbf{X}$,” where $\mathbf{C}$ represents a specific combination of semantic categories, and $\mathbf{X}$ denotes arbitrary contextual information that characterizes the surrounding environment.}

\subsection{IELFormer for Enhancing Domain Generalization} \label{sec:3.3 IELFormer}

\begin{figure*}[!t]
\centering
\includegraphics[width=6.8in]{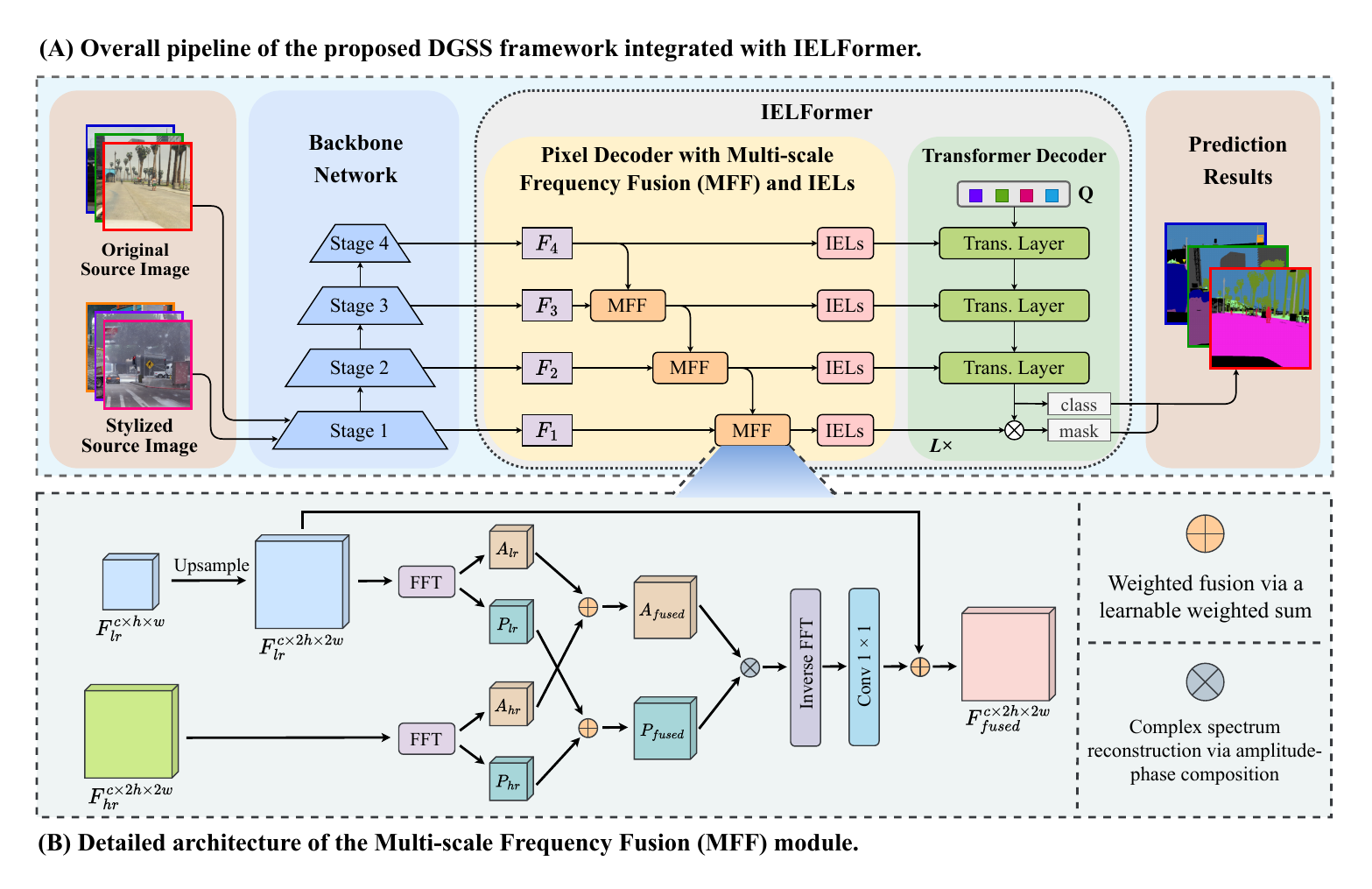}
\caption{\textbf{(A) Overall pipeline of the proposed DGSS framework integrated with IELFormer.} This framework progressively refines features through frequency fusion and inverse evolution mechanisms to improve generalization ability across diverse domains. \textbf{(B) Detailed architecture of the Multi-scale Frequency Fusion (MFF) module.} The module adaptively fuses amplitude and phase components from multi-resolution features to capture complementary information across scales.}
\label{figure:IELFormer}
\end{figure*}

An overview of the proposed DGSS framework is depicted in Fig. \ref{figure:IELFormer}, outlining the complete processing pipeline. Specifically, we alternate between feeding the original source images and their high-quality stylized counterparts generated by the proposed IELDM into a shared backbone network, enabling semantic feature extraction under diverse style representations. These features are then decoded using a Mask2Former-based \cite{cheng2022masked} segmentation head to produce dense predictions. To further enhance segmentation accuracy, we introduce IELs into the decoding stage. Moreover, we observe that the original Mask2Former lacks an effective mechanism for multi-scale feature aggregation, which compromises its ability to fully leverage the rich hierarchical semantic representations embedded in intermediate feature maps. To address this limitation, we propose a \textbf{multi-scale frequency fusion (MFF)} strategy, where features from different resolutions are transformed into the frequency domain to better capture and integrate complementary information across scales. By integrating the IELs with our frequency-aware multi-scale fusion design, we construct a novel decoder architecture termed \textbf{IELFormer}, which serves as an effective extension of the Mask2Former framework for DGSS.
\\

\noindent \textbf{IELs for Mitigating Prediction Noise in DGSS.} Most existing efforts in DGSS have primarily focused on designing various strategies such as domain-invariant representation learning or style diversification to directly enhance the generalization capacity of the model across unseen domains. While these methods have demonstrated significant progress, they may still face an inherent challenge: the predictions generated by the model, especially under strong domain shifts, are often imperfect to some extent. When such less-than-ideal predictions are repeatedly fed back into the training loop, there is a risk that semantic noise may accumulate, potentially reinforcing incorrect patterns and affecting the model’s ability to generalize effectively. To fundamentally address this challenge, we propose incorporating IELs into the DGSS framework. Motivated by their demonstrated effectiveness in the previously introduced IELDM, IELs are explicitly designed to highlight and enhance the negative components within predicted segmentation maps, thereby improving the robustness of the overall model.

As illustrated in Fig. \ref{figure:IELFormer}, IELs are integrated at four feature scales within the pixel decoder, each placed immediately after the corresponding intermediate-resolution feature map. Embedding IELs at multiple depths enables the network to progressively rectify prediction artifacts that emerge at different scales, forming a hierarchical enhancement pathway for spatial and semantic refinement. By selectively accentuating these defects, IELs deliver explicit and targeted gradient feedback, directing the model’s attention toward semantically ambiguous or structurally inconsistent regions. This iterative correction process not only improves semantic fidelity and structural precision but also strengthens the model’s resilience to distributional shifts, enhancing intra-class consistency and fostering the learning of more transferable representations. Consequently, IELs contribute to a more coherent final segmentation and a stable, generalizable optimization process across diverse target domains.
\\

\noindent \textbf{Multi-Scale Frequency Fusion (MFF).} To effectively integrate multi-scale features and enhance structural consistency across resolutions, we propose a multi-scale frequency fusion (MFF) module, as illustrated in Fig. \ref{figure:IELFormer}. Given a low-resolution feature map $F_{lr} \in \mathbb{R}^{c \times h \times w}$ and a high-resolution counterpart $F_{hr} \in \mathbb{R}^{c \times 2h \times 2w}$, the MFF first upsamples the low-resolution feature to match the spatial resolution of the high-resolution counterpart, yielding $F_{lr}^{\uparrow} \in \mathbb{R}^{c \times 2h \times 2w}$. Subsequently, both $F_{lr}^{\uparrow}$ and $F_{hr}$ are transformed into the frequency domain using the 2D fast Fourier transform (FFT):

\begin{equation}\label{eq:FFT}
\hat{F}_{l r} = \mathrm{FFT}(F_{l r}^{\uparrow}), \quad \hat{F}_{h r} = \mathrm{FFT}\left(F_{h r}\right).
\end{equation}
Each transformed feature map is then decomposed into its amplitude and phase components:

\begin{equation}\label{eq:AmpPhase}
\hat{F} = A \cdot e^{jP},
\end{equation}
where $A = |\hat{F}|$, and $P = \arg(\hat{F})$. We denote the amplitude and phase of the upsampled low-resolution and high-resolution features as $A_{lr}, P_{lr}$ and $A_{hr}, P_{hr}$, respectively. Specifically, the amplitude encodes the structural intensity and low-frequency semantics of the image, while the phase captures high-frequency details and fine spatial alignment. By explicitly separating these two complementary aspects, we enable a more targeted fusion strategy for multi-scale features. These components are then fused independently via a learnable weighted sum:

\begin{equation}\label{eq:dot}
\begin{array}{cc}
A_{\text{fused}} = \alpha \cdot A_{\text{lr}} + (1 - \alpha) \cdot A_{\text{hr}}, \\
P_{\text{fused}} = \beta \cdot P_{\text{lr}} + (1 - \beta) \cdot P_{\text{hr}},
\end{array}
\end{equation}
where $\alpha$, $\beta \in [0, 1]$ are learnable parameters that control the contribution of each scale. Next, the fused amplitude and phase are recombined to form the fused frequency representation:

\begin{equation}\label{eq:spectrum}
F_{\text{freq}} = A_{\text{fused}} \cdot e^{j P_{\text{fused}}}.
\end{equation}

This complex spectrum is transformed back to the spatial domain using the inverse FFT (IFFT):

\begin{equation}\label{eq:IFFT}
F_{\text{ifft}} = \operatorname{IFFT}(\hat{F}_{\text{fused}}).
\end{equation}

Finally, to further refine the output and facilitate residual learning, a $1 \times 1$ convolution followed by an element-wise addition is applied:

\begin{equation}\label{eq:out}
F_{\text{fused}} = F_{l r}^{\uparrow} + \operatorname{Conv}_{1 \times 1}\left(F_{\text{fused}}\right).
\end{equation}

By decoupling amplitude and phase, the model independently modulates structural information (phase) and intensity information (amplitude), enabling more precise and flexible integration of complementary features across scales. This decoupling allows MFF to capture fine-grained textures while preserving global contextual patterns, thereby improving both spatial alignment and semantic richness in the fused representations.

%%%%%%%%%%%%%%%%%%%%%%%%%%%%%%%%%%%%%%%%%%%%%%%%%%%%%%%%%%%%%%%%%%%%%%%%%%%%%%%%%%%%%%%%%%%
%%%%%%%%%%%%%%%%%%%%%%%%%%%%%%%%%%%%%%%%%%%%%%%%%%%%%%%%%%%%%%%%%%%%%%%%%%%%%%%%%%%%%%%%%%%

\section{Experiments} \label{sec:experiments}

\subsection{Experimental Setups}\label{sec:4.1 Implementation Details}

\noindent \textbf{Datasets:} To assess the model’s capacity for generalization to new environments, we train on three source synthetic domains (GTA \cite{richter2016playing}, SYNTHIA \cite{ros2016synthia} and UrbanSyn \cite{gomez2025all}) and test on three unseen real-world domains (Cityscapes \cite{cordts2016cityscapes}, BDD100K \cite{yu2020bdd100k}, and Mapillary \cite{neuhold2017mapillary}). Specifically, the synthetic source domains include GTA, comprising 24,966 densely annotated images extracted from the GTAV game engine with a resolution of 1914 $\times$ 1052; SYNTHIA, which contains 9,400 photo-realistically rendered images at a resolution of 1280 $\times$ 760; and UrbanSyn, consisting of 7,539 images with a resolution of 2048 $\times$ 1024. For real-world target domains, Cityscapes provides 2,975 training and 500 validation images, all at a resolution of 2048 $\times$ 1024. Additionally, BDD100K and Mapillary contribute 1,000 and 2,000 validation images with resolutions of 1280 $\times$ 720 and 1902 $\times$ 1080, respectively. For conciseness, we refer to Cityscapes as \emph{City}, BDD100K as \emph{BDD}, and Mapillary as \emph{Map} throughout the remainder of this paper. In addition, we further augment the training set by incorporating 5,000 synthetic images generated by our proposed IELDM framework. These images are alternately mixed with the original source domain samples during training to ensure balanced exposure to both authentic and style-diversified data.
\\

\noindent \textbf{Implementation details:} For \textbf{DGSS}, we base our implementation on the MMSegmentation \cite{mmseg2020} framework, incorporating the widely recognized Mask2Former \cite{cheng2022masked} segmentation head with DINO-v2 \cite{oquab2023dinov2} as the backbone to ensure high-quality performance. During training, we adopt the AdamW\cite{loshchilov2017decoupled} optimizer, with the learning rate configured to 1e-5 for the backbone and 1e-4 for the decoding head. To ensure an efficient training process, we conduct 40,000 iterations with a batch size of 4, employing cropped image patches of resolution 512 $\times$ 512. Following the protocol in Rein \cite{wei2024stronger}, we adopt only basic data augmentation techniques throughout the training phase. For \textbf{IELDM}, we adopt Stable Diffusion v1.5 \cite{rombach2022high} as the foundational generative model due to its strong capacity in producing high-quality and semantically consistent images. To enhance controllability and customization in the generation process, we further integrate ControlNet \cite{zhang2023adding} to incorporate structural guidance, and employ DreamBooth \cite{ruiz2023dreambooth} to enable fine-grained personalization through subject-specific tuning following \cite{jia2024dginstyle}.
\\

\noindent \textbf{Evaluation metric:} For the evaluation of DGSS, we adopt mean Intersection over Union (\textbf{mIoU}) as the primary performance metric. In the image generation component, we employ CLIP-based Maximum Mean Discrepancy (\textbf{CMMD}) \cite{jayasumana2024rethinking} to evaluate the visual quality and distributional consistency of the generated images, which is particularly suitable for small-scale datasets and offers high computational efficiency. In addition, \textbf{mIoU} is utilized to assess the semantic alignment between the generated images and their corresponding ground-truth labels, reflecting how well the generated content preserves class-specific structures.

\subsection{Comparison with State-of-the-art DGSS Methods}\label{sec:4.2Comparisons with the state-of-the-arts}

\noindent \textbf{Quantitative Performance Evaluation.} To comprehensively evaluate the effectiveness of our proposed IELDG framework, we conduct extensive comparisons against a range of state-of-the-art DGSS methods under diverse synthetic-to-real generalization scenarios. As summarized in Tab. \ref{tab:sotaComparison}, IELDG consistently enhances the performance of two strong and recently proposed baselines, Rein \cite{wei2024stronger} and SoMA \cite{yun2025soma}, across all domain generalization settings. When trained solely on GTA and evaluated on the three unseen target domains, IELDG achieves improvements in the average mIoU over the three targets by up to $1.6\%$ compared with Rein and $1.2\%$ compared with SoMA, respectively. Comparable gains are also observed under the other two training configurations, underscoring the generalizability of our method. These results highlight the efficacy of incorporating IELs, which help rectify structural distortions and semantic inconsistencies in feature representations, thereby improving the robustness of learned features. Notably, such performance gains are attained with only a marginal increase in trainable parameters, indicating that our framework introduces minimal computational overhead.

To comprehensively evaluate the capability of our method in mitigating domain shift, we report the class-wise IoU over 19 semantic categories in Tab. \ref{tab:gta2csbddmap} for three representative domain generalization settings: GTA → City, GTA → BDD, and GTA → Map. Across all scenarios, the proposed IELDG framework exhibits consistently superior performance in the majority of categories. Compared to the baseline models, these gains are particularly pronounced in structurally complex and domain-sensitive categories such as \emph{rider}, \emph{motorbike}, and \emph{traffic light}, where IELDG achieves substantial improvements. These findings further underscore the synergistic advantages of the proposed framework. The IELDM-generated images substantially enrich the source domain with diverse structural variations, effectively alleviating overfitting to domain-specific artifacts. In parallel, IELs strengthen structural invariance and semantic consistency across domains. Through the joint contribution of these components, IELDG achieves significant gains in cross-domain generalization without incurring additional annotation costs, offering a practical and scalable solution for real-world deployment.

\begin{table}[!t]
  \centering
  \caption{Comparison of the proposed IELDG with existing DGSS methods under various synthetic-to-real settings. Our results are averaged over 3 random seeds. * denotes the number of trainable parameters within the backbone architectures, while $\dag$ indicates re-evaluated performance using official checkpoints to ensure methodological consistency and fair benchmarking.}
  \definecolor{verylightgray}{gray}{0.90}
  \resizebox{\textwidth}{!}{
    \begin{tabular}{lcccccc}
    \toprule
    \multicolumn{3}{c}{\textbf{Synthetic-to-Real Generalization}} & \multicolumn{4}{c}{Test Domains(mIoU)} \\
\cmidrule{4-7}    Methods & Backbone & Params.* & City  & BDD   & Map   & Avg. \\
    \midrule
    \multicolumn{7}{c}{\textbf{Single-source DGSS Trained on GTA}} \\
    CLOUDS\cite{benigmim2024collaborating} {\scriptsize\textcolor{gray}{[CVPR'24]}} & CLIP-CN-L & 0.0M  & 60.20  & 57.40  & 67.00  & 61.50  \\
    VLTSeg\cite{hummer2024vltseg} {\scriptsize\textcolor{gray}{[ACCV'24]}} & EVA02-L & 304.2M & 65.30  & 58.30  & 66.00  & 63.20  \\
    DoRA\cite{liu2024dora} {\scriptsize\textcolor{gray}{[ICML'24]}} & DINOv2-L & 7.5M  & 66.12 & 59.31 & 67.07 & 64.17 \\
    Rein\cite{wei2024stronger} {\scriptsize\textcolor{gray}{[CVPR'24]}} & DINOv2-L & 3.0M  & 66.40  & 60.40  & 66.10  & 64.30  \\
    VPT\cite{jia2022visual} {\scriptsize\textcolor{gray}{[ECCV'22]}} & DINOv2-L & 3.7M  & 68.75 & 58.64 & 68.32 & 65.24 \\
    tqdm\cite{pak2024textual} {\scriptsize\textcolor{gray}{[ECCV'24]}} & EVA02-L & 304.2M & 68.88 & 59.18 & 70.1  & 66.05 \\
    FADA\cite{bi2024learning} {\scriptsize\textcolor{gray}{[NeurIPS'24]}} & DINOv2-L & 11.7M & 68.23  & 61.94  & 68.09  & 66.09  \\
    AdaptFormer\cite{chen2022adaptformer} {\scriptsize\textcolor{gray}{[NeurIPS'22]}} & DINOv2-L & 6.3M  & 70.10  & 59.81  & 68.77  & 66.23  \\
    SSF\cite{lian2022scaling} {\scriptsize\textcolor{gray}{[NeurIPS'22]}} & DINOv2-L & 0.5M  & 68.97  & 61.30  & 68.77  & 66.35  \\
    LoRA\cite{hu2022lora} {\scriptsize\textcolor{gray}{[ICLR'22]}} & DINOv2-L & 7.3M  & 70.13  & 60.13  & 70.42  & 66.89  \\
    SoMA\cite{yun2025soma} {\scriptsize\textcolor{gray}{[CVPR'25]}} & DINOv2-L & 4.9M  & 71.82  & 61.31  & 71.67  & 68.27  \\
    \midrule
    Rein$\dag$\cite{wei2024stronger} {\scriptsize\textcolor{gray}{[CVPR'24]}} & DINOv2-L & 3.0M  & 67.18  & 60.16  & 65.95  & 64.43  \\
    \rowcolor{verylightgray}Rein + IELDG (Ours) & DINOv2-L & 3.8M  & 68.76  & 60.51  & 68.90  & 66.06  \\
    SoMA$\dag$\cite{yun2025soma} {\scriptsize\textcolor{gray}{[CVPR'25]}} & DINOv2-L & 4.9M  & 70.35  & 61.20  & 70.49  & 67.34  \\
    \rowcolor{verylightgray}SoMA + IELDG (Ours) & DINOv2-L & 5.7M  & \textbf{72.18}  & \textbf{61.71}  & \textbf{71.66}  & \textbf{68.52}  \\
    \midrule
    \multicolumn{7}{c}{\textbf{Multi-source DGSS Trained on GTA + SYNTHIA }} \\
    Rein$\dag$\cite{wei2024stronger} {\scriptsize\textcolor{gray}{[CVPR'24]}} & DINOv2-L & 3.0M  & 69.54  & 61.12  & 67.38  & 66.01  \\
    \rowcolor{verylightgray}Rein + IELDG (Ours) & DINOv2-L & 3.8M  & 70.83  & 60.96  & 69.54  & 67.11  \\
    SoMA$\dag$\cite{yun2025soma} {\scriptsize\textcolor{gray}{[CVPR'25]}} & DINOv2-L & 4.9M  & 72.15 & 62.36 & 71.55 & 68.69 \\
    \rowcolor{verylightgray}SoMA + IELDG (Ours) & DINOv2-L & 5.7M  & \textbf{73.27} & \textbf{62.44} & \textbf{72.96} & \textbf{69.56} \\
    \midrule
    \multicolumn{7}{c}{\textbf{Multi-source DGSS Trained on GTA + SYNTHIA + UrbanSyn}} \\
    Rein$\dag$\cite{wei2024stronger} {\scriptsize\textcolor{gray}{[CVPR'24]}} & DINOv2-L & 3.0M  & 72.36 & 61.82 & 70.35 & 68.17 \\
    \rowcolor{verylightgray}Rein + IELDG (Ours) & DINOv2-L & 3.8M  & 73.89 & 62.12 & 70.76 & 68.92 \\
    SoMA$\dag$\cite{yun2025soma} {\scriptsize\textcolor{gray}{[CVPR'25]}} & DINOv2-L & 4.9M  & 74.58 & \textbf{62.43} & 73.92 & 70.31 \\
    \rowcolor{verylightgray}SoMA + IELDG (Ours) & DINOv2-L & 5.7M  & \textbf{75.89} & 61.63 & \textbf{74.97} & \textbf{70.83} \\
    \bottomrule
    \end{tabular}%
    }
  \label{tab:sotaComparison}%
\end{table}%

\begin{figure*}[!t]
\centering
\includegraphics[width=6.5in]{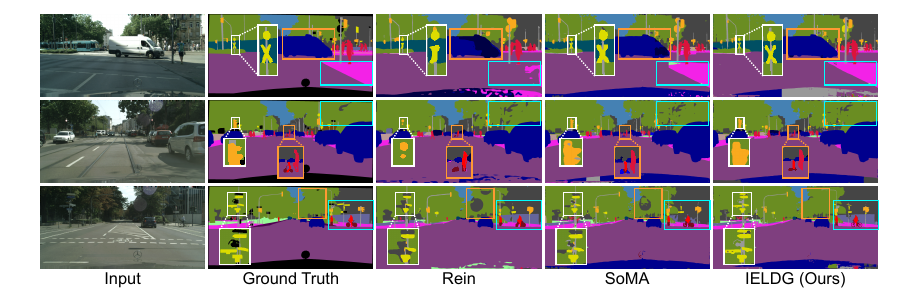}
\caption{Qualitative comparison of segmentation results on GTA → Cityscapes. Our proposed IELDG framework demonstrates superior visual accuracy and coherence compared to the two state-of-the-art baselines, Rein and SoMA, effectively preserving fine details and reducing artifacts in complex scenes.}
\label{figure:DGSSpred}
\end{figure*}

\noindent \textbf{Qualitative Analysis of Visual Results.} To further assess the visual effectiveness of the proposed IELDG framework, we present qualitative comparisons against two state-of-the-art baseline methods Rein \cite{wei2024stronger} and SoMA \cite{yun2025soma} on the GTA → City setting. As illustrated in Fig. \ref{figure:DGSSpred}, IELDG consistently yields more accurate and coherent segmentation results, particularly in challenging regions with fine-grained structures and occlusions. For instance, in the first row, IELDG correctly segments the sidewalk and the vehicle contours with higher boundary fidelity, while both Rein and SoMA suffer from category confusion and spatial disconnection. Similarly, in the second row, our method exhibits superior preservation of traffic lights, demonstrating robustness under domain shifts and low visibility, as well as maintaining accurate recognition of distant objects. These observations validate the effectiveness of our design in mitigating semantic degradation and improving the generalization capability of the segmentation model.

\begin{table*}[!t]
  \centering
  \caption{Quantitative comparison of IELDG with baseline methods Rein\cite{wei2024stronger} and SoMA\cite{yun2025soma} across three synthetic-to-real scenarios: GTA $\rightarrow$ City, GTA $\rightarrow$ BDD, and GTA $\rightarrow$ Map. Our results are averaged over 3 random seeds. mIoU are averaged over 19 categories, respectively. The best results are highlighted in \textbf{bold}. $\dag$ indicates re-evaluated performance using official checkpoints to ensure methodological consistency and fair benchmarking.}
  \footnotesize
  \definecolor{verylightgray}{gray}{0.90}
   \resizebox{\textwidth}{!}{\begin{tabular}{c|ccccccccccccccccccc|c}
    \toprule
    Methods & Road  & S.walk & Build. & Wall  & Fence & Pole  & Tr.light & Sign  & Veget. & Terrain & Sky   & Person & Rider & Car   & Truck & Bus   & Train & M.bike & Bike  & mIoU \\
    \midrule
    \multicolumn{21}{c}{GTA $\rightarrow$ City}\\
    \midrule
    Rein$\dag$\cite{wei2024stronger} {\scriptsize\textcolor{gray}{[CVPR'24]}} & 91.4  & 55.2  & 90.6  & 58.3  & 53.4  & 50.7  & 59.4  & 50.7  & 89.7  & 48.4  & 91.0  & 74.8  & 45.8  & 92.9  & 72.1  & 83.4  & \textbf{71.2}  & 42.5  & 54.6  & 67.2  \\
    \rowcolor{verylightgray}Rein + IELDG (Ours) & 93.5  & 60.1  & 91.8  & 58.8  & 56.2  & 51.7  & 68.8  & 55.4  & \textbf{91.1}  & \textbf{48.6}  & \textbf{91.9}  & 75.7  & 47.5  & 94.0  & 68.9  & 82.6  & 70.9  & 43.2  & 55.6  & 68.8  \\
    SoMA$\dag$\cite{yun2025soma} {\scriptsize\textcolor{gray}{[CVPR'25]}} & 93.6  & 63.2  & 91.6  & 56.9  & 54.6  & 59.2  & 67.7  & 55.8  & \textbf{91.1}  & 45.8  & \textbf{91.9}  & 80.1  & 47.3  & 94.0  & \textbf{73.4}  & \textbf{86.4}  & 70.6  & 54.0  & \textbf{59.4}  & 70.4  \\
    \rowcolor{verylightgray}SoMA + IELDG (Ours) & \textbf{94.7}  & \textbf{67.3}  & \textbf{92.0}  & \textbf{60.0}  & \textbf{59.7}  & \textbf{63.9}  & \textbf{69.2}  & \textbf{56.6}  & 90.9  & 46.8  & 91.8  & \textbf{82.2}  & \textbf{60.8}  & \textbf{94.9}  & 73.1  & 85.8  & 67.9  & \textbf{56.8}  & 57.1  & \textbf{72.2}  \\
    \midrule
    \multicolumn{21}{c}{GTA $\rightarrow$ BDD}\\
    \midrule
    Rein$\dag$\cite{wei2024stronger} {\scriptsize\textcolor{gray}{[CVPR'24]}} & 95.2  & 66.0  & 85.8  & 27.1  & 46.3  & 56.7  & 56.2  & 47.5  & 78.3  & 45.9  & 89.6  & 67.9  & \textbf{51.2}  & 90.4  & 56.1  & \textbf{74.7}  & 4.0  & 54.5  & 48.9  & 60.2  \\
    \rowcolor{verylightgray}Rein + IELDG (Ours) & \textbf{95.3}  & \textbf{67.3}  & 86.2  & 31.3  & \textbf{48.3}  & 55.5  & 59.3  & 50.5  & \textbf{79.4}  & 48.1  &  89.8  & 70.1  & 49.1  & 89.9  & 53.1  & 70.4  & 1.9  & 63.7  & 40.1  & 60.5  \\
    SoMA$\dag$\cite{yun2025soma} {\scriptsize\textcolor{gray}{[CVPR'25]}} & 95.0  & 64.7  & \textbf{86.5}  & \textbf{31.7}  & 47.9  & 57.2  & 58.2  & 47.1  & 78.8  & 47.6  & 89.4  & 70.9  & 44.5  & 90.9  & \textbf{59.4}  & 64.6  & \textbf{8.6}  & 68.1  & \textbf{50.9}  & 61.2  \\
    \rowcolor{verylightgray}SoMA + IELDG (Ours) & \textbf{95.3}  & \textbf{67.3}  & 85.5  & 28.2  & 47.3  & \textbf{57.8}  & \textbf{62.3}  & \textbf{50.7}  & 78.3  & \textbf{48.3}  & \textbf{90.1}  & \textbf{71.9}  & 44.2  & \textbf{91.3}  & 58.0  & 72.1  & 6.2  & \textbf{68.2}  & 49.3  & \textbf{61.7}  \\
    \midrule
    \multicolumn{21}{c}{GTA $\rightarrow$ Map}\\
    \midrule
    Rein$\dag$\cite{wei2024stronger} {\scriptsize\textcolor{gray}{[CVPR'24]}} & 92.5  & 67.6  & 86.5  & 50.4  & 53.3  & 50.5  & 56.0  & 51.1  & 80.5  & 50.8  & 94.5  & 70.2  & 53.4  & 88.1  & 62.2  & 81.3  & 57.5  & 59.4  & 46.9  & 66.0   \\
    \rowcolor{verylightgray}Rein + IELDG (Ours) & 93.1  & 68.8  & 88.5  & \textbf{51.1} & \textbf{56.0} & 57.5  & 68.3  & \textbf{59.8} & \textbf{82.4} & 54.0  & \textbf{94.8} & 77.4  & 47.7  & 90.3  & 68.0  & 82.4  & 50.8  & 67.4  & 50.8  & 68.9  \\
    SoMA$\dag$\cite{yun2025soma} {\scriptsize\textcolor{gray}{[CVPR'25]}} & 93.0  & 68.7  & \textbf{88.6} & 50.4  & 55.2  & 58.6  & 67.4  & 57.7  & 82.1  & 53.7  & 94.6  & 79.3  & 57.3  & 89.6  & 65.0  & 83.4  & \textbf{66.7} & 72.1  & 55.9  & 70.5  \\
    \rowcolor{verylightgray}SoMA + IELDG (Ours) & \textbf{93.4} & \textbf{70.4} & 88.1  & 50.4  & 55.4  & \textbf{60.1} & \textbf{70.0} & 59.4  & 82.2  & \textbf{54.4} & 94.6  & \textbf{81.0} & \textbf{69.7} & \textbf{91.3} & \textbf{69.4} & \textbf{83.7} & 52.5  & \textbf{76.9} & \textbf{58.5} & \textbf{71.6}  \\
    \bottomrule
    \end{tabular}}%
  \label{tab:gta2csbddmap}%
\end{table*}%

\subsection{Ablation Study on the Number of IELs}\label{sec:4.3 Ablation Study on the Number of IELs}

To explore the impact of IEL depth on model performance, we conduct a series of ablation experiments by varying the number of IELs in both IELDM and IELFormer (to isolate the effect of IEL depth, we omit the use of the MFF module in this ablation, ensuring that performance changes are solely attributed to the number of IELs). For IELDM, we evaluate the effect of varying IEL depth on both CMMD and mIoU, and provide corresponding visual results to illustrate the changes in image quality. For IELFormer, we report the average mIoU across three target datasets under different IEL depths in the single-source DGSS setting trained on GTA, highlighting how the number of IELs influences segmentation performance across domains.

\begin{table}[!t]
  \centering
  \caption{Generation Quality Comparison of IELDM Under Varying IEL Depths.}
  \setlength{\tabcolsep}{5pt}
  \renewcommand{\arraystretch}{1.3}
    \begin{tabular}{ccccc}
    \toprule
    \multirow{2}[4]{*}{Metrics} & \multicolumn{4}{c}{IEL Depths} \\
\cmidrule{2-5}          & 20    & 10    & \cellcolor{gray!30}\textbf{5}     & 0 \\
    \midrule
    CMMD↓ & 1.469 ($\uparrow$0.234) & 0.936 ($\downarrow$0.299) & \cellcolor{gray!30}\textbf{0.782 ($\downarrow$0.453)} & 1.235 \\
    mIoU↑ & 64.38 ($\downarrow$1.29) & 66.13 ($\uparrow$0.46) & \cellcolor{gray!30}\textbf{68.32 ($\uparrow$2.65)} & 65.67 \\
    \bottomrule
    \end{tabular}%
  \label{tab:IELDGNUM}%
\end{table}%

\begin{table}[!t]
  \centering
  \caption{Generalization Performance Comparison of Single-source DGSS Under Varying IEL Depths.}
  \setlength{\tabcolsep}{3pt}
  \renewcommand{\arraystretch}{1.3}
    \begin{tabular}{ccccc}
    \toprule
    \multirow{2}[4]{*}{Methods} & \multicolumn{4}{c}{IEL Depths} \\
\cmidrule{2-5}          & 20    & 10    & \cellcolor{gray!30}\textbf{5}     & 0 \\
    \midrule
    Rein\cite{wei2024stronger} + IELs & 63.59 (-0.84) & 64.56 (+0.13) & \cellcolor{gray!30}\textbf{64.92 (+0.49)} & 64.43 \\
    SoMA\cite{yun2025soma} + IELs & 66.47 (-0.87) & 67.51 (+0.17) & \cellcolor{gray!30}\textbf{67.86 (+0.52)} & 67.34 \\
    \bottomrule
    \end{tabular}%
  \label{tab:IELFORMERNUM}%
\end{table}%

\noindent \textbf{Quantitative Comparison.} Tab. \ref{tab:IELDGNUM} presents the performance variations under different IEL depths in terms of CMMD and mIoU. For the CMMD metric, which reflects the distributional alignment between generated and real images, we observe that a depth of 20 leads to a noticeable increase in CMMD, indicating degraded performance. This can be attributed to the excessive amplification of structural irregularities by deep IEL stacks, which over-enhance minor defects and introduce noise, ultimately impairing image realism. As the depth decreases to 10 and further to 5, the CMMD value steadily declines, reaching the lowest score of 0.782 at depth 5, suggesting that this setting achieves the most effective defect correction while avoiding overfitting and maintaining high visual fidelity. A similar trend is observed in the mIoU metric, which reflects the semantic consistency of the generated images. When the IEL depth is set to 20, the model produces a relatively low mIoU of 64.38, suggesting that excessive depth may lead to overcorrection and semantic distortion due to amplified noise. As the depth decreases, the semantic structure of the generated images becomes more coherent. Notably, at a depth of 5, the model achieves the highest mIoU of 68.32, indicating that a moderate IEL depth strikes a better balance between defect correction and semantic preservation.

Tab. \ref{tab:IELFORMERNUM} presents the average mIoU results on the GTA→\{City, BDD, Map\} benchmark under varying IEL depths within the IELFormer framework. Consistent with the trend observed in IELDM, integrating IELs with a depth of 20 results in a noticeable decline in generalization performance for both Rein and SoMA baselines. This suggests that overly deep IEL stacks may excessively amplify structural noise and disrupt the semantic integrity of the learned representations. The optimal performance is attained at an IEL depth of 5, where the mIoU improves by +0.49 for Rein and +0.52 for SoMA. This outcome indicates that a shallower IEL configuration effectively enhances domain-invariant features while avoiding over-smoothing or semantic distortion. Such a balance is likely due to the ability of moderate-depth IELs to emphasize salient structural cues without introducing redundant or misleading details, thereby fostering more robust and transferable feature representations across domains.

\begin{figure*}[!t]
\centering
\includegraphics[width=6.0in]{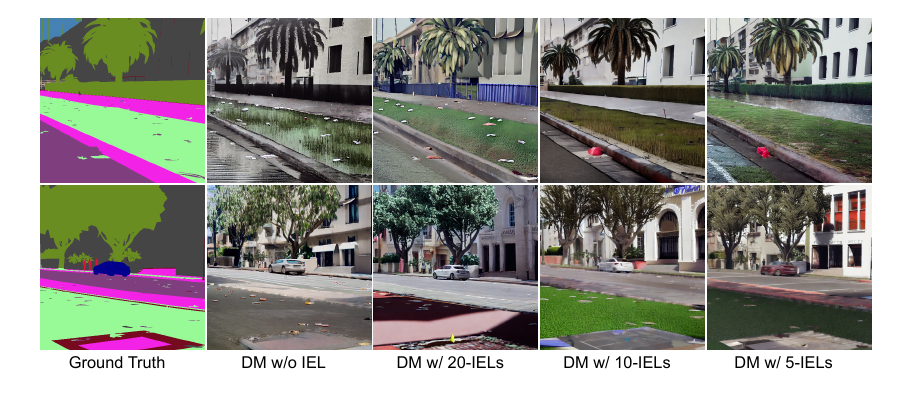}
\caption{Visual comparison of generated images under different IEL depths. Without IELs, the outputs suffer from severe artifacts and misclassification. As the IEL depth decreases, both structural fidelity and semantic accuracy improve, with the depth of 5 yielding the most realistic and semantically consistent results.}
\label{figure:IELNUM}
\end{figure*}

\begin{figure}[!t]
\centering
\includegraphics[width=3.5in]{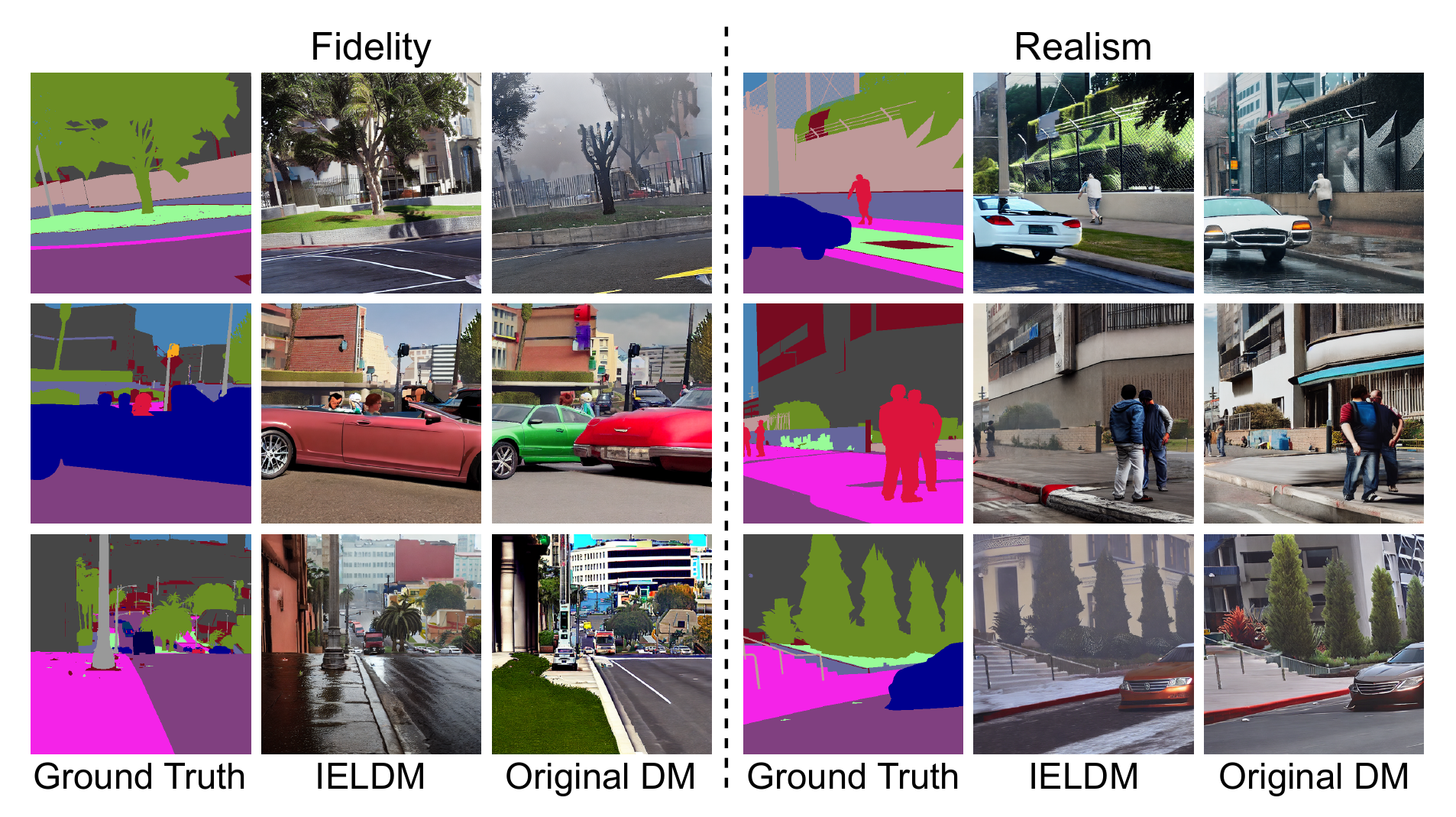}
\caption{Fidelity and realism comparison of generated images with and without IELs. Incorporating IELs leads to improved fidelity and enhanced realism.}
\label{figure:IELGEN}
\end{figure}

\noindent \textbf{Visual Analysis.} To provide more convincing evidence for the impact of IEL depth on the quality of generated images, we present visual comparisons under different configurations in Fig. \ref{figure:IELGEN}. To ensure a fair comparison, all image samples are generated under the same weather condition. When IELs are not applied, the generated images exhibit prominent visual flaws, including severe structural distortions and inaccurate category generation (e.g., the upper image misclassifies \emph{vegetation} as \emph{fence}, while the lower image misclassifies \emph{terrain} as \emph{sidewalk}), indicating poor semantic alignment. With an IEL depth of 20, the visual quality does not show meaningful improvement. A possible reason is that excessive depth amplifies subtle semantic defects, leading to the persistence of artifacts and misclassification of object categories. When the depth is reduced to 10, the generated images demonstrate fewer defects and improved semantic consistency, suggesting more effective suppression of noise and better structural preservation. The best visual quality is observed at a depth of 5, where images appear both semantically coherent and visually realistic. These visual trends align well with the quantitative results in Tab. \ref{tab:IELDGNUM}, further confirming that a moderate IEL depth yields the most balanced enhancement of image fidelity.

Moreover, the images generated by our proposed IELDM exhibit significantly higher fidelity and realism, as illustrated in Fig. \ref{figure:IELGEN}. In terms of fidelity, when IELs are not applied, the generated vegetation appears incomplete, a single vehicle is mistakenly rendered as two, and the base of a pole is incorrectly recognized as a car. These structural distortions are notably alleviated when IELs are incorporated, leading to more accurate and coherent reconstructions of scene elements. Regarding realism, the absence of IELs results in visually implausible artifacts such as blurry fences, merged humans, and stairs that lack distinct steps or proper structural detail. In contrast, incorporating IELs significantly enhances visual plausibility by producing clearer object boundaries and more coherent geometric structures.

\subsection{Ablation Study on IELDG Components}\label{sec:4.4 Ablation Study on IELDG Components}

\begin{table}[!t]
  \centering
  \caption{Ablation study of IELDG components on GTA→\{City, BDD, Map\}.}
  \definecolor{verylightgray}{gray}{0.90}
  \setlength{\tabcolsep}{4pt}
  \renewcommand{\arraystretch}{1.5}
    \begin{tabular}{cccccc}
    \toprule
    \multirow{2}[4]{*}{Methods} & \multirow{2}[4]{*}{Params*} & \multirow{2}[4]{*}{IELDM} & \multicolumn{2}{c}{IELFormer} & \multirow{2}[4]{*}{mIoU (Avg.)} \\
\cmidrule{4-5}          &       &       & MFF   & IELs   &  \\
    \midrule
    Rein\cite{wei2024stronger}  & 3.0M  & --     & --     & --     & 64.43 \\
    +IELDM & 0M    & $\checkmark$ & --     & --     & 65.36 (↑0.93) \\
    +MFF   & 0.8M  & --     & $\checkmark$ & --     & 64.78 (↑0.35) \\
    +IEL   & 0M    & --     & --     & $\checkmark$ & 64.92 (↑0.49) \\
    +MFF +IEL   & 0.8M    & --     & $\checkmark$ & $\checkmark$ & 65.13 (↑0.70) \\
    \rowcolor{verylightgray}+IELDM +IELFormer & 3.8M  & $\checkmark$ & $\checkmark$ & $\checkmark$ & \textbf{66.06 (↑1.63)} \\
    \midrule
    SoMA\cite{yun2025soma}  & 4.9M  & --     & --     & --     & 67.34 \\
    +IELDM & 0M    & $\checkmark$ & --     & --     & 68.12 (↑0.78) \\
    +MFF   & 0.8M  & --     & $\checkmark$ & --     & 67.62 (↑0.28) \\
    +IEL   & 0M    & --     & --     & $\checkmark$ & 67.86 (↑0.52) \\
    +MFF +IEL   & 0.8M    & --     & $\checkmark$ & $\checkmark$ & 68.06 (↑0.72) \\
    \rowcolor{verylightgray}+IELDM +IELFormer & 5.7M  & $\checkmark$ & $\checkmark$ & $\checkmark$ & \textbf{68.52 (↑1.18)} \\
    \bottomrule
    \end{tabular}%
  \label{tab:ablation}%
\end{table}%

\begin{figure*}[!t]
\centering
\includegraphics[width=5.8in]{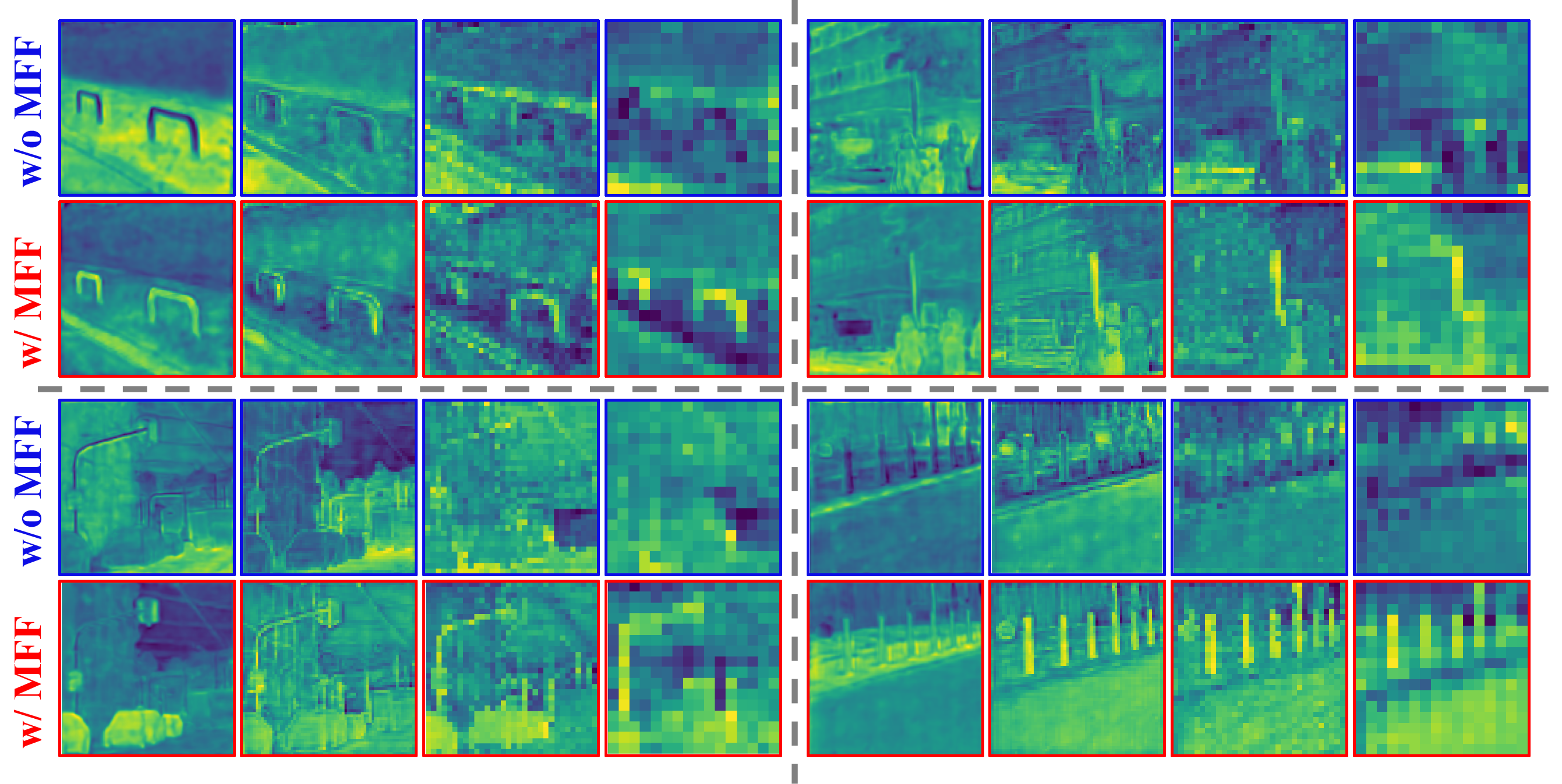}
\caption{Visual comparison of multi-scale feature maps with (red boxes) and without (blue boxes) the proposed MFF module. The inclusion of MFF enhances semantic coherence and structural consistency across layers, confirming its effectiveness in facilitating cross-scale feature integration.}
\label{figure:IELMFF}
\end{figure*}

To verify the effectiveness of each component in our proposed IELDG framework, we conduct ablation studies on the GTA→\{City, BDD, Map\} DGSS task. We report the average mIoU across the three target domains as the evaluation metric and validate the performance under two representative baselines: Rein \cite{wei2024stronger} and SoMA \cite{yun2025soma}. As summarized in Tab. \ref{tab:ablation}, when adopting Rein as the baseline, introducing high-quality images generated by IELDM into the training set yields a notable improvement of +0.93 mIoU. This performance gain is achieved without introducing additional trainable parameters, since the employed IELs are based on a fixed Laplacian kernel and the diffusion model used for generation remains frozen during the entire process. Furthermore, we analyze the contributions of the two major components of our IELFormer, namely multi-scale frequency fusion (MFF) and the IELs. When MFF alone is added, a gain of +0.35 mIoU is observed, indicating that capturing frequency-domain representations at multiple scales helps enhance semantic discrimination. When solely applying the IEL module, the improvement reaches +0.49 mIoU, demonstrating that explicitly attending to error-prone regions based on Laplacian-based priors can effectively guide the network to focus on spatial discontinuities and structural inconsistencies. When both components of IELFormer are integrated, the performance boost climbs to +0.70 mIoU, highlighting the complementarity and synergy between MFF and IELs. Ultimately, with the full IELDG pipeline, we achieve the best performance improvement of +1.63 mIoU, validating the cumulative effectiveness of all components. Similar trends are consistently observed when using SoMA as the baseline, further confirming the robustness and general applicability of our method across different architectures.

\subsection{Efficiency Analysis of MFF}\label{sec:4.5 Efficiency Analysis of MFF}

To enhance the integration of multi-scale features, we propose the multi-scale frequency fusion (MFF) module. This module adaptively recalibrates frequency components across different spatial resolutions, leveraging the complementary strengths of low- and high-frequency information to facilitate more effective cross-scale interaction and mitigate semantic inconsistencies across layers. To visually demonstrate the effectiveness of MFF, Fig. \ref{figure:IELMFF} presents a comparative analysis of multi-scale features with (red boxes) and without (blue boxes) the MFF module. For each group, features from left to right are derived from increasingly deeper layers, resulting in reduced spatial resolutions. Without MFF, the features exhibit scattered and incoherent patterns, especially in deeper layers, leading to diminished semantic clarity. In contrast, the features generated with MFF exhibit more coherent structures and improved semantic alignment across scales. These results affirm that MFF significantly improves the quality of feature representations, thereby contributing to more robust and accurate semantic understanding.

\section{Conclusion}
\label{sec:conclusions}
In this paper, we present a unified framework, IELDG, for domain generalized semantic segmentation, which simultaneously enhances the quality of synthetic training data and strengthens the robustness of segmentation models under domain shift. To address the limitations of diffusion-generated data that often contain semantic and structural defects, we propose IELDM, which integrates inverse evolution layers (IELs) into the generative process. IELDM effectively amplifies structural inconsistencies via Laplacian priors, enabling more precise correction of flawed generative patterns and producing images with higher semantic fidelity and visual realism. Building upon this idea, we further embed IELs into the segmentation network to form IELFormer, which suppresses prediction artifacts by highlighting uncertain regions at multiple semantic scales. To improve semantic coherence across feature hierarchies, we also introduce a multi-scale frequency fusion (MFF) module that performs frequency-domain decomposition and structured integration of multi-resolution features. Extensive experiments across multiple cross-domain benchmarks validate the effectiveness of our IELDG, demonstrating its potential as a robust and scalable solution for real-world deployment in domain-shifted environments.

\section*{Acknowledgments}
This work was supported by the Science and Technology Innovation Team Project of Shaanxi Province under Grant No. 2025RS-CXTD-013, the National Natural Science Foundation of China under Grant Nos. 11971379, and the Distinguished Youth Foundation of Shaanxi Province under Grant No. 2022JC-01. Z. Qiao's work is partially supported by the Hong Kong Research Grants Council RFS grant RFS2021-5S03, GRF grant 15305624 and NSFC/RGC Joint Research Scheme grant N\_PolyU5145/24.

%\section*{References}
\bibliographystyle{unsrt}
\biboptions{numbers,sort&compress}
\bibliography{reference}

\end{document}